\documentclass{article}

\usepackage{arxiv}
\usepackage{amsmath}
\usepackage{array}
\usepackage{graphics}
\usepackage{url,hyperref,lineno,microtype,subcaption}
\usepackage[onehalfspacing]{setspace}

\title{BENDR: using transformers and a contrastive self-supervised learning task to learn from massive amounts of EEG data.}

\author{
  Demetres Kostas\\
  University of Toronto, Toronto, Canada\\
  Vector Institute, Toronto, Canada \\
  \texttt{demetres@cs.toronto.edu} \\
  \And
  St\'{e}phane Aroca-Ouellette \\
  University of Toronto, Toronto, Canada\\
  Vector Institute, Toronto, Canada \\
  \And
 Frank Rudzicz \\
  University of Toronto, Toronto, Canada\\
  Vector Institute, Toronto, Canada \\
  Li Ka Shing Knowledge Institute, Toronto, Canada \\
}
\date{}
\renewcommand{\shorttitle}{BENDR}
\renewcommand{\undertitle}{}

\begin{document}

\maketitle

\begin{abstract}

Deep neural networks (DNNs) used for brain-computer-interface (BCI) classification are commonly expected to learn general features when trained across a variety of contexts, such that these features could be fine-tuned to specific contexts. While some success is found in such an approach, we suggest that this interpretation is limited and an alternative would better leverage the newly (publicly) available massive EEG datasets. We consider how to adapt techniques and architectures used for language modelling (LM), that appear capable of ingesting awesome amounts of data, towards the development of encephalography modelling (EM) with DNNs in the same vein. We specifically adapt an approach effectively used for automatic speech recognition, which similarly (to LMs) uses a self-supervised training objective to learn compressed representations of raw data signals. After adaptation to EEG, we find that a single pre-trained model is capable of modelling completely novel raw EEG sequences recorded with differing hardware, and different subjects performing different tasks. Furthermore, both the internal representations of this model and the entire architecture can be fine-tuned to a \emph{variety} of downstream BCI and EEG classification tasks, outperforming prior work in more \emph{task-specific} (sleep stage classification) self-supervision.

\end{abstract}

\section{Introduction}

To classify raw electroencephalography (EEG) using deep neural networks (DNNs), discriminative models need to both extract useful features from raw sequences, and classify those features. This frames both the promise and the challenge of using DNNs: feature engineering could be almost entirely avoided, without introducing limitations on classifier complexity, but both feature extraction and classification need to be learned from a \emph{limited} supply of (relevant) high-dimensional data. This challenge is evident in brain-computer interface (BCI) applications, where DNNs can struggle to determine good features. A large degree of data variability within and between different users causes the classification performance of many model types to vary \cite{Sannelli2019, Lotte2018, Dose2018, Ahn2015}. Fundamentally, this reveals that these models lack generality, and instead rely on characteristics specific to particular subjects (and/or sessions). Furthermore, beyond these inter- and intra-personal variations, different features are relevant for different BCI tasks in the first place. Hand-selected features (sets possibly pruned later on) are distinct under different BCI paradigms, as different features better discriminate different tasks\footnote{While this is typical, some procedures, like covariance-based Riemannian classification schemes, do not necessarily need different features for different tasks \cite{Lotte2018, Zanini2018}.} \cite{Lotte2018}, e.g., P300 versus motor imagery. In other words, unlike domains such as computer vision where there is a clearer understanding that nearly all DNNs tend to learn ``low-level'' features in earlier layers (e.g., edge-detector-like primitives) \cite{Raghu2019, Yosinski2015, Krizhevsky2012}, there is no such understanding with DNNs used to process raw EEG. There are no known transferable DNN properties or operations that are easily extended to any subject, session, or task. Importantly however, the determination of which ``low-level'' features DNNs developed in computer vision was revealed through models that had transferable performance from general to specific tasks\cite{Raghu2019, Yosinski2015}. The development of transferable DNNs for raw EEG then appears to be a promising classification tool on the one hand, but could also serve to validate existing techniques, and perhaps even suggest novel methods (if early layers do or do not correspond to existing methodologies respectively).

The difficulty of learning both ``lower-level'' features and an expressive classifier simultaneously may help explain why work using DNNs to classify raw BCI data has tended to prefer shallower networks \cite{Kostas2020, Roy2019, Lotte2018, Lawhern2018, Schirrmeister2017}. With these shallower networks, the range of \emph{learnable} features is relatively limited. By design, these employ constrained linear operations, and a limited few of these layers include subsequent non-linear activations \cite{Kostas2020}, an otherwise crucial feature of DNN complexity. Fundamentally, their inability to uniformly outperform feature-engineering approaches \cite{Lotte2018} indicate that these limited features are not entirely sufficient, and more importantly, they may not always be desirable in a DNN approach \cite{Kostas2020}. In prior work we presented evidence that, if inter-personal variability had been adjusted for, the performance of shallower models more quickly saturates to lower performance levels as compared to a deeper network alternative \cite{Kostas2020}, suggesting that more complex raw-BCI-trial features \emph{could} be developed using DNNs with sufficient data, notably such that these data provide a reasonable empirical estimate of the data distribution in question. Inter/intra-person variability sabotages this approach, limiting the inter-applicability of data from all people and sessions of an entire dataset. This is disappointing since the labelling process is much more difficult than in other domains of DNN research to begin with\footnote{Consider the difficulty of collecting and labelling 100 more BCI trials as compared to the same for 100 more images.}.

In this work, we argue that self-supervised sequence learning would be an effective approach for developing and deploying more complex DNNs in BCI, as it can learn from many more people, sessions, and tasks using \emph{unlabelled} data, thus promising to better model the input distribution of EEG data; it affords the possibility to learn features with little variability across traditionally confounding factors. Specifically, we investigate techniques inspired by language modelling (LM), that have found recent success in self-supervised end-to-end speech recognition and image recognition. We begin by comparing fully supervised transfer learning (which has been frequently looked to as an EEG/BCI TL solution) to self-supervised approaches, finding inconsistency in the extension of computer vision-style pre-training to BCI (and by extension the data domain of EEG). We then evaluate a simple adaptation of previous work in self-supervised speech recognition called \texttt{wav2vec 2.0}\cite{Baevski2020} to EEG. With this framework, arbitrary EEG segments are encoded as a sequence of learned feature vectors we call BErt-inspired Neural Data Representations (or `BENDR'). We ask whether BENDR are: transferable to novel EEG recorded from unseen subjects, different hardware, and different tasks, and if BENDR are generally suitable (both as-is or fine-tuned) to a battery of downstream EEG classification tasks. 

\subsection{Pre-training with DNNs}
\label{sec:prior}

For inspiration on tackling DNN pretraining in BCI, one can look to successful applications in other domains. The modern deep learning (DL) ``revolution'' was ushered in on the back of computer vision and image recognition \cite{Sejnowski2020, LeCun2015}. The successes of DL in this domain have stemmed from a lineage of massive \emph{labelled} datasets \cite{LeCun2015}, such as the ImageNet dataset \cite{imagenet}. These datasets were used to train deep convolutional neural networks, often one of the variants or progeny of ResNet \cite{He2015} and DenseNet \cite{Huang2016}. Crucially, these are labelled datasets, featuring -- especially in the case of ImageNet -- an enormous number of unique possible classification \emph{targets} (1000 is common with ImageNet\footnote{\url{image-net.org/challenges/LSVRC/2012/}}, but more are possible\footnote{\url{http://image-net.org/about-stats}}). As mentioned above, leveraging labelled data (especially for a particular task) of a similar scale in BCI is impractical but, despite this, a sizeable amount of prior work tries to fashion a transfer learning strategy after the successes of ImageNet pre-training. These take the form of transferring knowledge from a network trained with \emph{more data}, typically more subjects, to a target domain with \emph{less data}, typically a single subject \cite{Kostas2020, Fahimi2019, Xu2019, Dose2018, Schwemmer2018, Lin2017a}, with some work transferring between entire datasets of the same paradigm, rather than subjects \cite{Ditthapron2019}. On the surface, these embody a general-to-specific supervised transfer learning scheme reminiscent of ImageNet pre-training. However, these particular framings lack diversity in pre-training targets. Instead, the number and type of targets remains the same in both the pre-training and fine-tuning stages. We remain unaware of any work that pre-trains a DNN with a \emph{wide gamut of BCI-relevant targets} to a \emph{more narrow} target set, as would be common when using ImageNet as pre-training for more specific computer vision tasks\footnote{It is also worth noting that our own prior work does not consider or identify this.}. This is noteworthy, as this is part of what makes ImageNet a \emph{general task}. Evidence suggests that pre-training label diversity is important for effective ImageNet transfer learning \cite{Huh2016}, though an excess could be detrimental \cite{Ngiam2018, Huh2016}. More fundamentally, however, this pre-training paradigm has begun to be questioned altogether, with some work finding that it does not necessarily improve downstream performance, where commonly it has been assumed that it should (e.g., in medical images or object localization; though it \emph{speeds up} training considerably) \cite{He2019a, Raghu2019, Kornblith2019, Ngiam2018}.

What has begun to emerge as a potential alternative in computer vision -- and markedly so when there is limited labelled downstream data -- is self-supervised learning \cite{Chen, Grill2020, Henaff2019a, Oord2018} \footnote{Terminology here can be somewhat fuzzy. What is meant by self-supervision is a supervision-like task that requires domain-relevant understanding in some sense. Sometimes, `semi-supervised' is used instead, as it is often also a semi-supervised procedure \cite{Chen}, since the task is learned in an unsupervised fashion first and then classic supervised learning is used with labels. Typically, though, semi-supervision involves inferring labels for unlabelled data during training. Instead, self-supervision is loosely a particular case of representation learning, which is not historically uncommon in BCI \cite{Zhang2020}. Though this work is different given that typically the loss is domain or data agnostic.}. These works are inspired by the recent success in natural language processing (NLP) using LMs, which can be used for transfer learning, but also for few-shot and zero-shot learning \cite{Brown2020, Raffel2019a}. We propose that DNN transfer learning in BCI and neuroimaging analysis generally could follow a similar line, with \emph{encephalography models} (EM) in place of LMs. The important question being {\em how best to construct such an EM, so that it learns features that are general enough while remaining usable for any analysis task?}

Prior work has developed approaches for (EEG) self-supervised sleep stage classification (SSC) through contrastive learning\cite{Banville2019a}. Contrastive learning in its most general form consists of identifying positive representations from a set that also includes incorrect or negative distractor representations \cite{Arora2019}. Banville \emph{et al.} proposed two potential contrastive learning tasks -- a ``relative positioning'' task and an extension they termed ``temporal shuffling''\cite{Banville2019a}. Underlying both tasks is the notion that neighbouring representations share a label. This is a fair assumption for SSC, where sleep stages change slowly, and is generally reasonable for continuous problems, where some notion of smoothness is assumed. Their proposed ``relative positioning'' task is a binary classification problem distinguishing whether a pair of representations are within a local or positive window $\tau_{pos}$, or outside a long-range or negative window $\tau_{neg}$ (when $\tau_{neg} > \tau_{pos}$, those falling within $\tau_{neg}$ but outside $\tau_{pos}$ are ignored). The representations themselves are a learned mapping (in their case, a convolutional neural network) of raw EEG time-windows to a feature vector. Their alternative ``temporal shuffling'' method adds a third window or representation with which to contrast that is within $\tau_{pos}$ of one (arbitrary) window called the `anchor', and again learns the representations through a binary classification task. In this case, the classification determines whether the three representations are ordered sequentially, or are out of order. These tasks ultimately both improved downstream SSC performance over the same network trained in a fully supervised manner with randomly initialized weights (with self-supervision being distinctly better when limiting fine-tuning data, a common theme in the recent wave of self-supervision literature \cite{Chen2020a, Brown2020}) and a variant of the same network but trained under an autoencoder paradigm (alternative pretraining option; the network was pretrained to reconstruct the original waveform). ``Relative positioning'' performed better on average (and no statistical significance expressed) as compared to its counterpart, but a linear classification of simple hand-crafted features was still highest performing overall. These results demonstrate the promise of self-supervised learning with DNNs for EEG over a supervised approach. This is all the more valuable, as it appeared that the self-supervised time-window representations (learned features of a window), when projected into a 2D visualization, also appeared to model some sleep-stage information and information about subject age \cite{Banville2019a} purely through the contrastive task without use of sleep-stage labels. The major concern with these particular schemes though are the lengths of the time windows ($\tau_{pos}$ and $\tau_{neg}$). The shortest windows employed were 2 minutes for $\tau_{pos}$ and $\tau_{neg}$, which seems prohibitively long. As it is assumed that representations within $\tau_{pos}$ are similarly labelled, it may be difficult to expand the use of this technique to time scales closer to that of a BCI trial (across any paradigm), which tend to be no more than several seconds at most. Instead, we focus our efforts on adapting a relevant strategy from the wider ML literature that could develop features on smaller time scales.

Returning to transfer learning successes in NLP, the \emph{masked} language model (MLM) is a slight variation on the typical LM which models the probability of encountering a language token given previous (or, in some cases, also subsequent) tokens. The MLM scheme instead learns to \emph{reconstruct} language token(s) given surrounding context (fashioned after the Cloze task), and is employed by BERT \cite{Devlin2018} and its lineage \cite{Raffel2019a} of similar models, which embody part of the recent wave of successful NLP transfer learning. This family of models may deploy a variety of auxiliary tasks \cite{Aroca-Ouellette2020} for transfer learning, but the task currently at the heart of this family is as follows: given a sequence of $N$ tokens $t_1, ... t_N$, and a subset of token indexes $I_m$, for each token index $i \in I_m$, tokens are masked with some mask $M$ so that:

\begin{equation}
\label{eq:mlm}
    q_i = \begin{cases}
            M; i \in I_m \\
            t_i; \text{otherwise}
          \end{cases}, \forall i \in N     
\end{equation}

A transformer encoder \cite{Devlin2018, Vaswani2017} then reconstructs the original sequence of tokens from the \emph{masked} sequence ($t_i$ and $q_i, \forall i \in N$ respectively in eq. \ref{eq:mlm}). $M$ could be a single learned token \cite{Baevski2020}, or in the case of BERT: 80\% of the time a fixed \texttt{[MASK]} token, 10\% a random token or 10\% the original token (with 15\% of tokens masked within each sequence) \cite{Devlin2018}. 

Could an EM be developed in this vein, using individual samples rather than tokens (i.e., direct application of BERT to raw EEG)? Unfortunately, the highly correlated nature of neighbouring samples in EEG (or most other continuous data for that matter), is not conducive to this approach. The likely result would be that, instead of an EM, a method for interpolation would be learned, as has been argued in similar work in self-supervised learning with speech \cite{Jiang}. In other words, the smoothness of these data would make it hard to produce general features simply through recovering missing points. Masking a contiguous span of tokens instead, which is beneficial in NLP \cite{Joshi2020, Raffel2019a}, could avoid simply learning to interpolate missing samples, but the \emph{reconstruction} of time-series data is difficult, due to the difficulty (among other things) of capturing the degree of error in time (within contiguous sequences) \cite{Rivest2020}. The losses used for such reconstruction, commonly mean squared error (or mean absolute error), erroneously assume independence in the error between elements in the series, causing inappropriate error signals when (among other things) when simply shifting a reconstruction in time \cite{Rivest2020}. 

Contrastive predictive coding (CPC), is a contrastive learning-based task that retains the character of sequence learning provided by masked language model-like approaches, but is not as susceptible to degeneration into interpolation, or similarly affected by the issues of time-series reconstruction \cite{Oord2018}. With CPC, the correct \emph{learned representation} for a particular sequence offset is predicted relative to distractor representations, typically those of other positions in the same sequence \cite{Oord2018}. This task enables learning both a good feature representation and an understanding of the progression of those features end-to-end. Interestingly, both the representations alone \cite{Chen2020a}, and the addition of the sequence model \cite{Baevski2020} have proven potentially useful for supervised fine-tuning after pre-training.

Prior work in self-supervised speech recognition has begun to synthesize parts of CPC and MLM to produce methodologies for self-learning with raw waveforms \cite{Baevski2020, Baevski2020a, Chung, Jiang, Oord2018}. In our work, we adapt one of these approaches called \texttt{wav2vec 2.0} \cite{Baevski2020} (its particular formulation is detailed in section \ref{sec:pretraining}) to EEG, and investigate how effective the representations (BENDR) are for downstream tasks.

\section{Materials and methods}

All experiments are implemented using the \emph{deep neural networks for neurophyisiology} (DN3) library \footnote{https://github.com/SPOClab-ca/dn3}. The source code and pre-trained BENDR models can be found at \url{https://github.com/SPOClab-ca/BENDR}. 

\subsection{Datasets}

The ideal pre-training dataset for our purposes would feature many subjects, each recorded over many sessions. These sessions would also ideally be distributed across large time-scales and consist of a variety of performed tasks. In other words, the pre-training dataset should consist of a representative sample of EEG data in the most general sense. This also means that these data should include multiple different recording hardware and configurations. The closest publicly accessible dataset, to our current knowledge, was the Temple University Hospital EEG Corpus (TUEG) \cite{tueg}. It consists of clinical recordings using a mostly conventional recording configuration (monopolar electrodes in a 10-20 configuration) of over 10,000 people, some with recording sessions separated by as much as eight months apart. The subjects were 51\% female, and ages range from under 1 years old to over 90 \cite{tueg}. We focused specifically on versions 1.1 and 1.2 of this dataset which amounted to approximately 1.5 TB of European-data-format (EDF) EEG recordings \emph{before} preprocessing. 

Furthermore, we compiled a non-exhaustive battery of publicly accessible EEG data classification tasks summarized in table \ref{tab:ds-summaries}. Most of these were BCI task datasets, which could readily be compared to previous work with DNNs trained without any additional unlabelled data \cite{Kostas2020, Lawhern2018}. We also included one of the sleep stage classification (SSC) tasks used by Banville \emph{et al.} \cite{Banville2019a} in their work on sleep stage self-supervision described above, for comparison. This dataset afforded some further insight into generality, as BCI data are typically classified in the context of particular trials or events, and SSC is a more continuous problem, requiring that large spans of time are labelled with the particular sleep stage a subject is undergoing. These segments are distinctly longer than the BCI trials we considered in the remaining battery (an order of magnitude difference in our case when compared to the largest BCI task sequence length), and are distinctly closer in length to the pre-training task. This allowed us to consider how effective our approach was to such a different time-scale. Another notable difference with the SSC dataset was the scale of available labels, which seems to have enabled prior work to consider deeper and more complex models \cite{Mousavi2019}. We segmented these sequences into 30 second periods as in prior work, and focused on 5 labels as in prior work \cite{Mousavi2019, Banville2019a}.

\begin{table}[h]
    \centering
    \setlength{\extrarowheight}{4pt}
    \begin{tabular}{l|c|c|c|c|c|c}
    \hline
        Dataset                                      & Paradigm                 & sfreq. H\emph{z} & \# Ch. & Subjects & Targets & Folds \\ \hline \hline
        MMI \cite{Schalk2004, physionet}             & MI (L/R)                 & 160              & 64     & 105      & 2       & 5     \\ \hline
        BCIC \cite{Tangermann2012}                   & MI (L/R/F/T)             & 250              & 22     & 9        & 4       & 9     \\ \hline
        ERN \cite{kaggleERN}                         & Error Related Negativity & 200              & 56     & 26 (10)  & 2       & 4     \\ \hline
        P300 \cite{erpbci-article,erpbci,physionet}  & Donchin Speller          & 2048             & 64     & 9        & 2       & 9     \\ \hline
        SSC \cite{sleep-edf, sleep-data, physionet}  & Sleep Staging            & 100              & 2      & 83       & 5       & 10    \\ \hline
        
    \end{tabular}
    \caption[Summary of downstream dataset battery and number of cross-validation folds used. Cross validation splits were in a leave-multiple-subjects-out configuration if $Folds < Subjects$, or leave-one-subject-out if $Folds = Subjects$ (as in prior work \cite{Kostas2020}). The ERN dataset was featured in an online competition \url{https://www.kaggle.com/c/inria-bci-challenge} which featured 10 held-out test subjects (not used during training), which we used as a test dataset for all four validation splits of this dataset.]{Summary of downstream dataset battery and number of cross-validation folds used. Cross validation splits were in a leave-multiple-subjects-out configuration if $Folds < Subjects$, or leave-one-subject-out if $Folds = Subjects$ (as in prior work \cite{Kostas2020}). The ERN dataset was featured in an online competition\footnotemark~which featured 10 held-out test subjects (not used during training), which we used as a test dataset for all four validation splits of this dataset.}
    \label{tab:ds-summaries}
\end{table}

\footnotetext{\url{https://www.kaggle.com/c/inria-bci-challenge}}

\subsection{Preprocessing}
\label{sec:preprocessing}

The focus of the preprocessing stage was to create a maximally consistent representation of EEG sequences across datasets, so that the pre-trained network was well-suited to a \emph{variety} of ``downstream'' tasks. More or less, this amounted to modifying downstream datasets to match the configuration of the pre-training dataset. The first aspect of this was to remove spurious differences in channel amplitude. Each sequence gathered for training was linearly scaled and shifted (a weight and offset for each sequence adjusts every sample in the sequence) so that the maximum and minimum values within each sequence equal $1$ and $-1$ respectively. To account for the lost relative (to the entire dataset) amplitude information, a single channel was added with the constant value $\frac{max(s_i) - min(s_i)}{max(S_{ds}) - min(S_{ds})}$, where $S_{ds}$ is the set of all samples in the dataset and $s_i \subset S_{ds}$ is a particular sub-sequence (i.e., trial). We additionally addressed the differences in sampling frequency and electrode sets of the different dataset. Our solutions to these problems were similarly minimalist and were achieved using standard features in DN3 \cite{dn3-tvectors}. Specifically, we over- or under-sampled (by whole multiples, for lower and higher sampling frequencies respectfully) to get nearest to the target sampling frequency of 256 H\emph{z}. Then, nearest-neighbour interpolation was used to obtain the precise frequency (described further in \cite{dn3-tvectors}). Additionally, the P300 dataset was low-pass filtered below 120 H\emph{z} to avoid aliasing due to its higher sampling rate (and associated higher original low pass filter). Furthermore, the SSC dataset featured two bi-polar electrodes: FPz-Cz and Pz-Oz, which were simply mapped to FPz and Pz, respectively. The TUEG dataset also features some higher sampling rate signals; we included those with low-pass filters that did not violate the Nyquist criterion (and subsequently re-sampled them as above), and ignored the rest.

A reduced subset of the \texttt{Deep1010} channel mapping from DN3 \cite{dn3-tvectors} was used throughout. This ensured that particular channels were mapped to a consistent index for each loaded trial. The original mapping was designed to be more inclusive, and thus assumed up to 77 possible EEG electrodes. In the interest of minimizing unnecessary electrodes for an already high-dimensional problem, we focused on the 19 EEG channels of the \emph{unambiguously illustrated 10/20} channel set (UI 10/20) \cite{Jurcak2007}, as the TUEG dataset recordings were done using a roughly 10/20 channel scheme. We simply ignored reference electrodes, electro-oculograms, and any other auxiliary channels. When also accounting for the additional relative amplitude channel described above, every sequence from every dataset used 20 channels. All surplus channels were ignored, and missing channels set to $0$.

During pre-training, we extracted sequences of 60 seconds (every 60 seconds) from each usable sequence, which amounted to $15,360$ samples per subsequence. We observed in early testing that there was better performance with larger sequences (see figure \ref{fig:seq-regression} for more). As can be seen in table \ref{tab:ds-perf}, the downstream datasets all used sequence lengths shorter than this, but the architecture we employed (see section \ref{sec:model}) was ostensibly agnostic to sequence length  (see section \ref{sec:discussion} for caveats).

\subsection{Model architecture}
\label{sec:model}

The model architecture closely follows that of \texttt{wav2vec 2.0} \cite{Baevski2020} and is comprised of two stages. A first stage takes raw data and dramatically downsamples it to a new sequence of vectors using a stack of short-receptive-field 1D convolutions. The product of this stage is what we call BENDR (specifically in our case, when trained with EEG). A second stage uses a transformer \emph{encoder} \cite{Vaswani2017} (layered, multi-head self-attention) to map BENDR to some new sequence that embodies the target task. 

Raw data is downsampled through the stride (number of skipped samples) of each convolution block in the first stage (rather than pooling, which would require greater memory requirements). Each of our convolution blocks comprised of the sequence: 1D convolution, GroupNorm \cite{Wu2020}, and GELU activation \cite{Hendrycks2016}. Our own encoder features six sequential blocks, each with a receptive fields of 2, except for the first, which was 3. Strides matched the length of the receptive field for each block. Thus, the \emph{effective sampling frequency} of BENDR is 96 times smaller ($\approx 2.67$ H\emph{z}) than the original sampling frequency ($256$ H\emph{z}). Each block consists of 512 filters, meaning each vector has a length of 512.

The transformer follows the standard implementation of Vaswani \emph{et. al} \cite{Vaswani2017}, but with internal batch normalization layers removed and with an accompanying weight initialization scheme known as T-Fixup \cite{Huang2020}. Our particular transformer architecture uses 8 layers, with 8 heads, model dimension of 1536 and an internal feed-forward dimension of 3076. As with \texttt{wav2vec 2.0}, we use GELU activations \cite{Hendrycks2016} in the transformer, and additionally include LayerDrop \cite{Fan2019} and Dropout at probabilities $0.01$ and $0.15$, respectively, during pre-training but neither during fine-tuning. We represent position using an additive (grouped) convolution layer \cite{Baevski2020, Mohamed2019} with a receptive field of 25 and 16 groups before the input to the transformer. This allows the entire architecture to be sequence-length independent, although it may come at the expense of not properly understanding position for short sequences.

Originally, the downstream target of the \texttt{wav2vec 2.0} process was a downstream speech recognition \emph{sequence} (it was fine-tuned on characters and phonemes) \cite{Baevski2020}. Instead, here the entire sequence is classified. To do this using a transformer, we adopt the common practice \cite{Devlin2018} of feeding a fixed token (\emph{a.k.a.} \texttt{[CLS]} in the case of BERT or, in our case, a vector filled with an arbitrary value distinct from the input signal range, in this case: $-5$) as the first sequence input (prepended to BENDR). The transformer output of this initial position was not modified during pre-training, and only used for downstream tasks.

The most fundamental differences in our work as compared to that of the speech-specific architecture that inspired it are: 1. we do not quantize BENDR for creating pre-training \emph{targets}, and 2.  we have \emph{many} incoming channels. In \texttt{wav2vec 2.0}, a \emph{single} channel of raw audio was used. While a good deal of evidence \cite{Kostas2020, Kostas2019, Lawhern2018, Chambon2018, Lotte2018, Schirrmeister2017} supports the advantage of temporally-focused stages (no EEG channel mixing) separate from a stage (or more) that integrates channels, we elected to preserve the 1D convolutions of the original work to minimize any additional confound and to reduce complexity (compute and memory utilization $\propto N_{filters}$ with 2D rather than $\propto \frac{N_{filters}}{N_{EEG}}$ for 1D convolutions). This seemed fair, as there is also evidence that 1D convolutions are effective feature extractors for EEG, particularly with large amounts of data \cite{Gemein2020a, dn3-tvectors}. Notably, \texttt{wav2vec 2.0} downsampled raw audio signals by a much larger factor (320) than our own scheme, but speech information is localized at much higher frequencies than encephalographic data is expected to be. The new effective sampling rate of BENDR is $\approx 2.67$ H\emph{z}, or a feature-window (no overlap) of $\approx 375 ~ms$. We selected this downsampling factor as it remained stable (i.e., it did not degenerate to an infinite loss, or simply memorize everything immediately) during training.

\subsection{Training}

We used the Adam \cite{Kingma2015} optimizer throughout training, with weight decay set to $0.01$. We additionally used a cosine learning rate decay with linear warmup for 5\% and 10\% of total training steps (batches) for pre-training and fine-tuning respectively. The peak learning rate itself varied by dataset; this, and other variable hyperparameters, are further documented in appendix \ref{app:hyperparams}.

\subsubsection{Pre-training}
\label{sec:pretraining}

The pre-training procedure largely follows \texttt{wav2vec 2.0}, but we make some notable hyperparameter changes. Specifically, the self-supervised loss for a masked token localized at BENDR position $t$, is defined as:

\begin{equation}
    \mathcal{L} = -log \frac{exp(cossim(c_t, b_t)) / \kappa}{\sum_{b_i \in B_D} exp(cossim(c_t, b_i)) / \kappa}
\end{equation}

Where $c_t$ is the output of the transformer at position $t$, $b_i$ is the BENDR vector at some offset $i$, and $B_D$ is a set of 20 uniformly selected distractors from the same sequence, plus $b_t$. We use the cosine similarity $cossim(x, y) = x^Ty / (|x| |y|)$ function to determine how similar vectors are, and the sensitivity of this is adjusted by a temperature factor $\kappa$, set to $0.1$. In essence, this loss operates by adjusting the output of the transformer at position $t$ to be \emph{most similar to the encoded representation at $t$, despite that this input to the transformer is masked}. We also added the mean squared activation of the BENDR to the loss, as was similarly done previously \cite{Baevski2020}, but set the weight of this additional term to 1 (rather than 10).

We learn a single mask vector during pre-training of the same length as each BENDR vector, and use this as the transformer input to masked positions. Contiguous sequences of 10 are masked with probability $p_{mask} = 0.065$, such that, for each sample, the likelihood of being the \emph{beginning} of a contiguous section was $p_{mask}$, and overlap is allowed. The number of negatives/distractors was set to 20 and uniformly sampled from the \emph{same} sequence as the masked vector, i.e., negatives do not cross trials or sequences.

After pre-training, we examined how generalizable the sequence model and vectors were to unseen data, by evaluating the contrastive task, expressed as the transformer accuracy in constructing $c_t$ to be most similar to $b_t$ rather than the distractors. During evaluation, we masked half the amount expected during training, but such that masked spans were evenly spaced through the sequence (so that there were no overlapping sequences, and sufficient context was available). That is, for a sequence length of $N_S$, we masked $0.5 \times N_S \times p_{mask} = N_m$ contiguous sequences (of 10), and spaced them every $\left \lfloor{\frac{N_S}{N_m}}\right \rfloor$ steps (starting at the first sample). $N_S$ first remained at $15,360$ (60 seconds as in training, no overlap between subsequent sequence representations) for all datasets except P300, where sessions were too short and instead $5120$ (20 seconds) was used. We then evaluated the change in performance across the downstream datasets, excluding P300, as $N_S$ varied from 20-60 seconds.

\subsubsection{Downstream fine-tuning}
\label{sec:finetuning}

Ultimately, our aims for subject-, session-, and dataset-generalizable representations were not simply to accurately mimic the correct input, but with the intent that these representations -- and potentially the sequence model itself -- could be effectively transferred to specific and arbitrary tasks. We considered six different variations of TL across the battery of EEG classification tasks presented in table \ref{tab:ds-summaries}:

\begin{enumerate}
    \item Add a new softmax classification layer to the first (pre-pended position) output token of the transformer and train the entire model to classify the downstream targets.
    \item Ignore the pre-trained transformer, average pool the BENDR to four concatenated vectors, add a new classification layer and train the model (only the first stage and new layer) to classify the downstream targets.
    \item The same as (1.), but without pre-training
    \item The same as (1.), but keep the BENDR fixed and continue training the transformer.
    \item The same as (2.), but without pre-training
    \item The same as (2.), but keep the first stage weights fixed and train only the new classification layer.
\end{enumerate}

We considered these permutations so that we could speak to the effect each stage had on downstream performance, at least to some degree. First, we were interested in 1) determining whether the new sequence representation (BENDR) contained valuable features \emph{as-is} (as they appear to  for speech \cite{Baevski2020}) or if they required some further training, and 2) whether the sequence model learned characteristics of the BENDR that were informative to the classification task. Finally, ignoring pre-training all-together, of course, was to examine how effective the network would be at learning the task otherwise, without pre-training or transfer learning.

The P300, ERN, and SSC datasets all had imbalanced class distributions; we adjusted for these imbalances by \emph{undersampling} points of the more frequent classes with replacement so that the number of samples drawn -- per epoch -- of each class was equal to the number of examples of the least frequent target class.

We also included the sequence regularization proposed by \texttt{wav2vec 2.0} \cite{Baevski2020}, though we adjusted it for our more varied trial lengths. That is, in all 6 fine-tuning configurations, contiguous sections of 10\% of the entire BENDR of a trial were masked with the mask token learned during pre-training (not changed after pre-training) at a probability of $0.01$. In other words, this was the likelihood of a sample being the beginning of a contiguous masked section, as in pre-training. Additionally across the BENDR (throughout each vector in the sequence), a similar procedure dropped features to 0, where contiguous sections of 10\% of the channels (51) were dropped with a probability of $0.005$.

\section{Results}

\subsection{Pre-training generalization}

\begin{figure}
    \centering
    \includegraphics{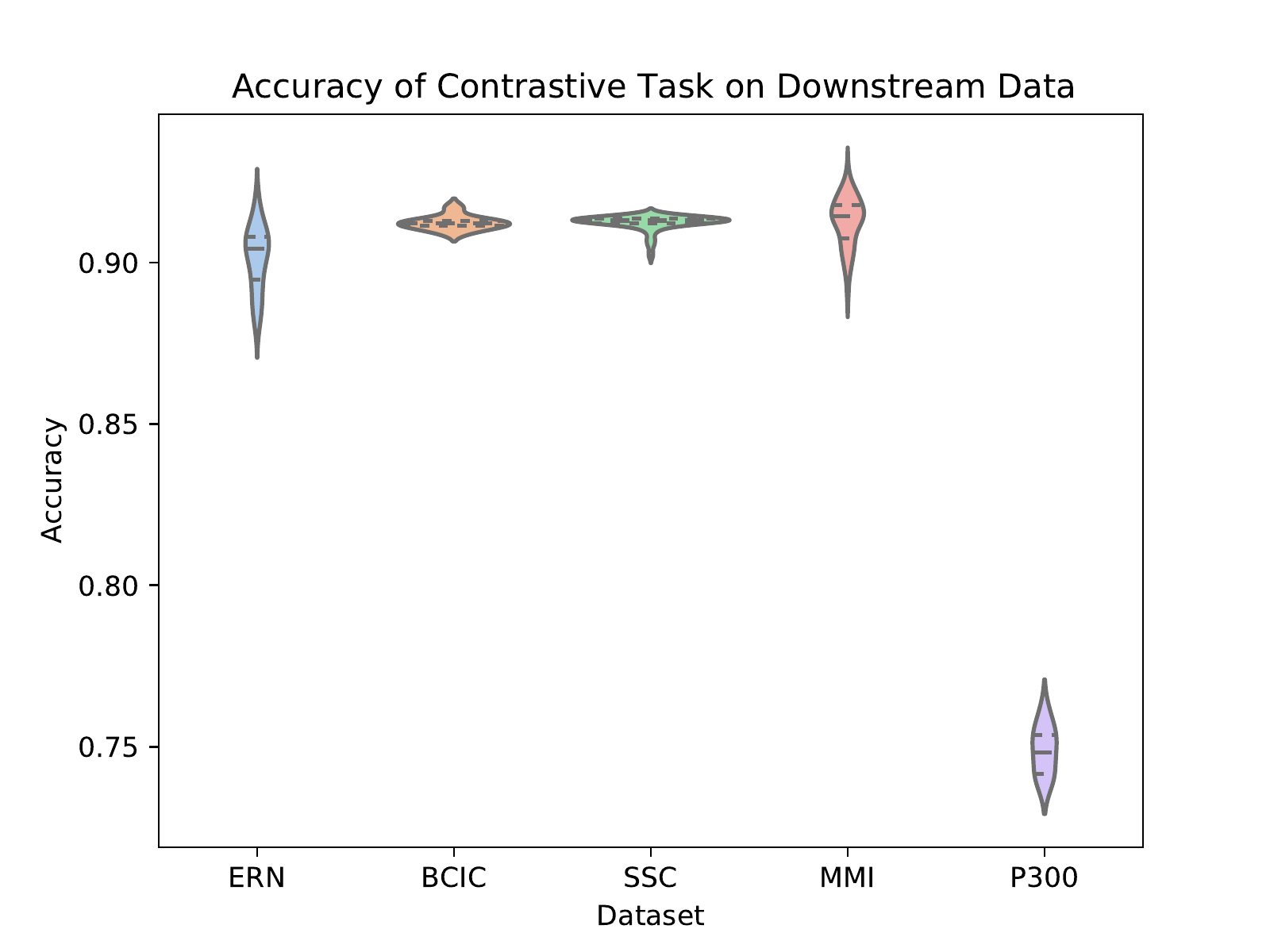}
    \caption{Violin plot (inner lines for quartile divisions) of test-subject-wise accuracy for each downstream dataset. Specifically, accuracy of the sequence model (transformer stage) at creating a representation that is closest to the correct representation at masked sequence positions. The P300 dataset is distinctly lower performing than the remaining datasets, though this was likely due to its shorter evaluation context (see figure \ref{fig:seq-regression}). Nonetheless, there is minimal test-subject-wise variation, particularly when compared to classifier performance generally.}
    \label{fig:sequences}
\end{figure}

\begin{figure}
    \centering
    \includegraphics{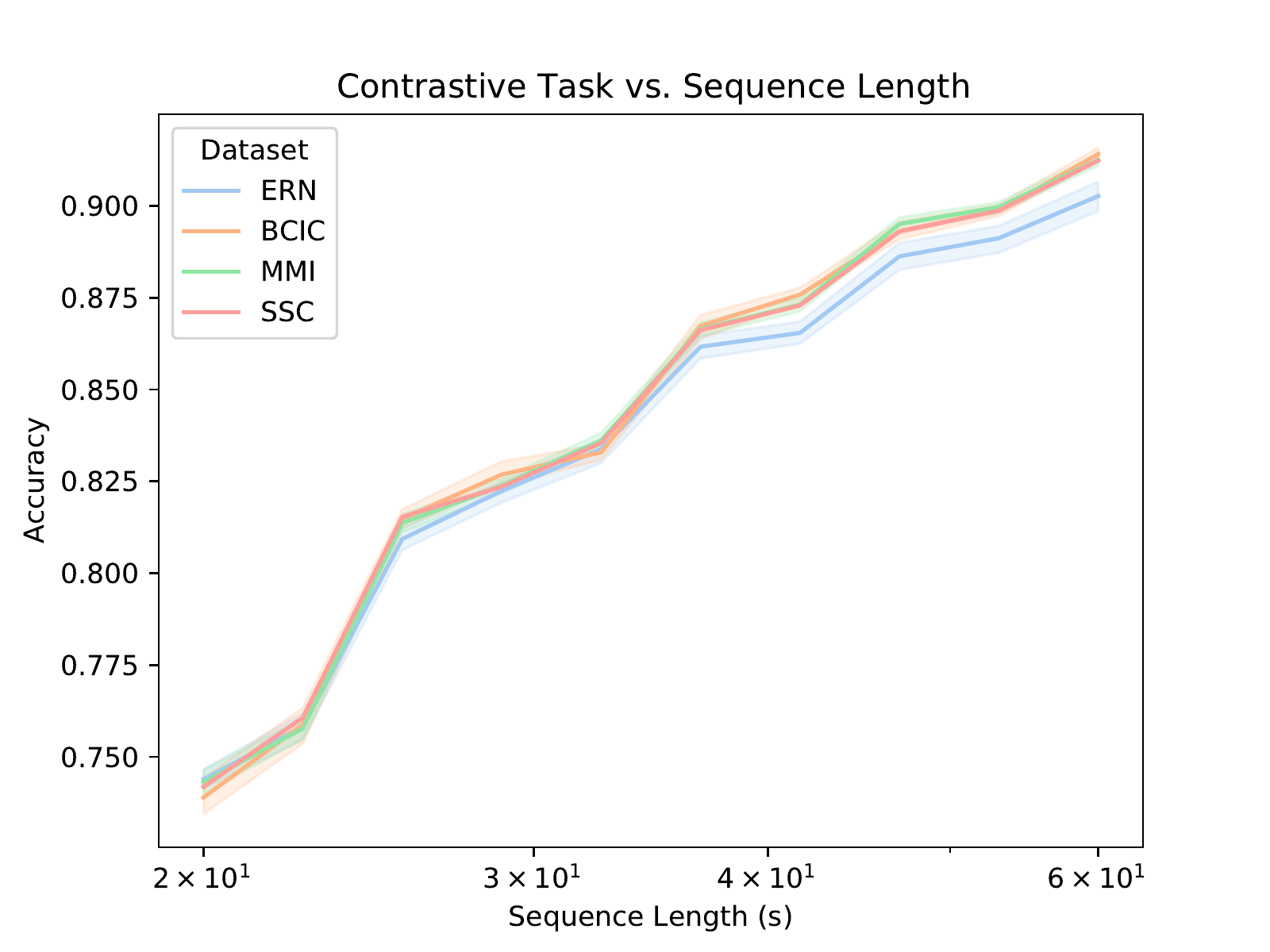}
    \caption{Contrastive accuracy versus evaluation length in seconds (x-axis logarithmic). Performance is distinctly similar for all datasets, rising for longer sequences. We suggest that this implies that samples that are further apart are easier to distinguish between than neighbouring samples. Thus, while BENDR encode local signal characteristics well, there is redundancy.}
    \label{fig:seq-regression}
\end{figure}

Figure \ref{fig:sequences} shows how accurate the transformer stage is at producing an appropriately similar BENDR when compared to distractor representations. There are two key observations in this figure, the first is that there is little variability across the first four datasets, and within each of the five datasets. The latter point implies that this accuracy is not radically variable across different subjects, as it tends to be when considering classifier performance \cite{Sannelli2019, Dose2018} (though, when fine-tuning for classification, this variability returns; see figure \ref{fig:downstream}). This could be because a) the transformer adequately learns a general model of how BENDR sequences of novel persons and equipment progressed, b) the BENDR themselves are invariant to different people, hardware, and tasks, c) some combination of the last two possibilities, or d) the problem is being solved via some non-signal characteristics. We return to this question shortly. The second observation was alluded to already: the P300 dataset distinctly under-performs the other downstream datasets. However, this coincided with the shortest evaluation sequence. Looking at figure \ref{fig:seq-regression}, we see that all five datasets have consistently similar performance when evaluated with 20 seconds of data, so the dip in P300 performance of figure \ref{fig:sequences} seems less remarkable. Taken together, \ref{fig:sequences} and \ref{fig:seq-regression} clearly indicate that a longer evaluation context makes the contrastive task easier. This suggests that the contrastive task is, in fact, solved by learning signal-relevant features, rather than some more crude solution like interpolation, or by simply creating a sequence of recognizable position representations (both of which have no reason to exhibit this dependence on sequence length). We believe the most likely explanation for the rise in performance with more context is that local representations are more difficult distractors, implying that the new effective sampling rate remains too high (and there is still redundant information encoded in local BENDR). Notwithstanding, there is a strong uniformity of performance across datasets and subjects (in both figures \ref{fig:sequences} and \ref{fig:seq-regression}), meaning this scheme develops features (whether through the transformer itself, or the BENDR) that generalize to novel subjects, hardware, and tasks, though their applicability to downstream contexts remains to be seen. 

\subsection{Downstream fine-tuning}

\begin{figure}
    \centering
    \includegraphics{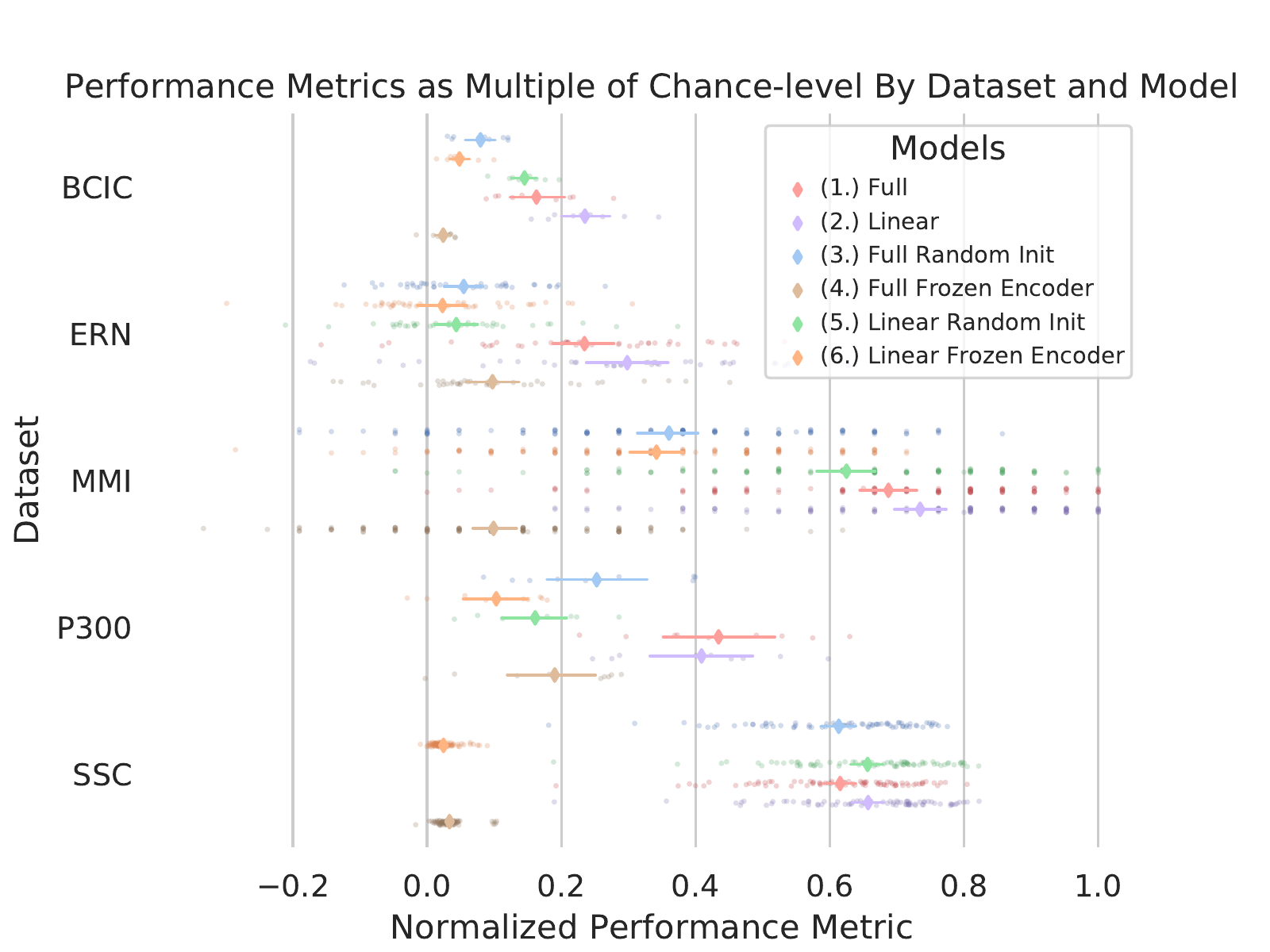}
    \caption{Performance of all downstream datasets for each of the six model configurations considered. Metrics vary by dataset, see table \ref{tab:ds-perf}. Metrics were normalized to range from chance (0) to perfect (1). Individual translucent points are performances of single subjects (within each test fold), solid diamonds indicate mean performance across all subjects/folds, with surrounding bars showing $.95$ confidence intervals using $n=1000$ bootstrap sampling. The discretized pattern of the MMI dataset is due to the limited trials \emph{per subject}, which resulted in limited distribution of performance levels. Notably here, (1.) or (2.) were consistently among the best performing, yet both remained within the confidence levels of each other. The randomly initialized average-pooled BENDR with linear classifier (5.) also performed well, though less consistently. Model configurations are numbered in accordance with the list presented in section \ref{sec:finetuning}.}
    \label{fig:downstream}
\end{figure}

\begin{table}
    \centering
    \setlength{\extrarowheight}{4pt}
    \begin{tabular}{|l|c|c|c|c|c|c|c|}
        \hline
        Dataset & Start (s) & Length (s) & Metric   & Best & Model config. \\ \hline \hline
        MMI     & 0         & 6          & BAC      & 86.7 & Linear (2.)    \\ \hline
        BCIC    & -2        & 6          & Accuracy & 42.6 & Linear (2.)   \\ \hline
        ERN     & -0.7      & 2          & AUROC    & 0.65 & Linear (2.)   \\ \hline
        SSC     & 0         & 30         & BAC      & 0.72 & Linear (2.)   \\ \hline
        P300    & -0.7      & 2          & AUROC    & 0.72 & BENDR (1.)    \\ \hline
    \end{tabular}
    \caption{Performances of downstream datasets. Start and length refer to length of trials and start with respect to event markers in seconds. Best performance specifies average performance across all subjects (and therefore folds) for best performing model configuration. BAC: class balanced accuracy; AUROC: area under the receiver operating characteristic curve. Model configurations are numbered in accordance with the list presented in section \ref{sec:finetuning}.}
    \label{tab:ds-perf}
\end{table}

Figure \ref{fig:downstream} and table \ref{tab:ds-perf} present a picture of how effectively BENDR could be adapted to specific tasks. Overall, the fine-tuned linear classification (listed as downstream configuration 2. above) that bypassed the transformer entirely after pre-training was highest performing four out of five times, though using the transformer for classification (1.) performed consistently similarly (confidence intervals always overlapped), and surpassed the bypassed transformer (2.) with the P300 dataset (and was highest performing for this dataset). Deploying the full network (initial stage and transformer) without pre-training was generally ineffective, though this was not the case with the SSC dataset, which may have been due to the larger data availability. In fact, for both the full and linear model architectures trained with the SSC data, fine-tuning the pre-trained model is mostly on par with the randomly initialized counterpart. Considering our results with the SSC data relative to those of Banville \emph{et. al}'s \cite{Banville2019a} proposed contrastive learning for sleep staging (described in section \ref{sec:prior}), their reported results show that the fine-tuned variants of our own model (1. and 2.) achieved a higher mean balanced accuracy relative to their two proposed schemes. Taken in concert with our own approach's wider applicability and more fine-grained temporal feature development, we believe this demonstrates that ours is a promising alternative. Interestingly, with and without pre-training (2. and 5.) achieved similar performance to Banville \emph{et. al}'s fully supervised results (where our configurations and their architecture employ similar 1D convolution-based schemes), which is notable as with this dataset, both their ``temporal-shuffling'' and ``relative-positioning'' tasks under-performed full supervision when utilizing the full SSC dataset.

Our fine-tuned approaches similarly appear reasonably competitive with prior work on the MMI dataset \cite{Kostas2020, Dose2018}, particularly when considering that only 19 channels (rather than the full set of 64) were being used. In all considered configurations, despite heavy regularization (and the very low learning rates) the randomly initialized parameters were consistently prone to overfitting, all the more so with the full model architecture. Conversely, the pre-trained networks were slow to fit to the downstream training data (under the exact same training scheme for fine-tuning). Ultimately, though most of these results are not necessarily state-of-the-art, this single pre-training scheme nonetheless shows a breadth of transferability which is apparently unique. 

\section{Discussion}
\label{sec:discussion}

We are unaware of any prior work assessing transformer-based \cite{Vaswani2017} DNNs with EEG data (raw or otherwise). This is perhaps consistent with the ineffectiveness we observed with the randomly initialized full architecture (3.) and could imply that effective use of this powerful emerging architecture \emph{requires} pre-training (or at least enough data, given the better looking SSC performance). Future work should continue to evaluate this architecture, particularly as it appears to be more widely applicable than the NLP applications it was originally proposed for \cite{Dosovitskiy2020, Baevski2020}.

We believe that our approach can be improved through adjusting the neural network architecture and pre-training configuration such that it becomes more data-domain (EEG) appropriate. Future work will prioritize effective integration of spatial information, likely by better isolating temporal and spatial operations. Evaluation using large downstream datasets that \emph{also} feature many channels, such as the {\em Montreal Archive of Sleep Studies} (MASS)\footnote{\url{http://massdb.herokuapp.com/en/}} will be considered. Though available for public access at the time of writing, these data were unavailable while experiments were prepared and conducted. Prior work shows that DNN approaches effective for EEG leverage spatial information \cite{Chambon2018}, and it is presently unclear to what degree this is the case with BENDR. In terms of data-appropriate temporal modelling, which we have considered with relatively more zeal in this work, recall that figure \ref{fig:seq-regression} presents the possibility that local representations may be retaining redundant information, further improvements may be found in better compressing the temporal resolution of BENDR. Future work will consider larger downsampling factors in the initial stage, along with longer sequences, balancing the more difficult problem of summarizing more data (in effect, further data \emph{compression}), with the apparent increased effectiveness of the contrastive task (as observed in figure \ref{fig:seq-regression}) on longer sequences. A small but potentially fruitful avenue for further improvement includes reconsidering the additive convolutional layer as a substitute for explicit position encodings, which are in fact more common \cite{Raffel2019a, Devlin2018, Vaswani2017}. Recall that this was originally for two reasons: \texttt{wav2vec 2.0} did the same, and we felt it best to limit excessive changes to the architecture on a first iteration, and because it seamlessly supported flexible input lengths. This latter point comes, however, with a trade-off -- our particular position encoder had a receptive field of 25 (stride of 1), which means a little over 9 seconds of input. While it seems that convolutional position encodings offer better performance \cite{Mohamed2019}, this input width exceeded the \emph{entire} length of all but the sleep classification task (the length we chose was optimized for pre-training behaviour).

After considering these possible avenues for improving BENDR, we still do not fully discount the validity of some of the transfer learning paths we appear to exclude above in our introduction. We will reconsider these paths in future work. Particularly, given the success we had in crossing boundaries of hardware in this work, and in prior work \cite{dn3-tvectors}, it may be possible to construct an \emph{aggregate} dataset featuring a variety of EEG classification tasks, towards better ImageNet-like pre-training. The construction of a more coherent label set that crosses several BCI paradigms would no doubt be a significant effort (e.g., problems may include: is a rest period before one task paradigm the same as rest before another? What about wakeful periods in sleep?). This would no doubt be imbalanced; the labels would be distributed in a long-tailed or Zipfian distribution that would likely require well thought-out adjustment \cite{Tang2020, Cao2019}. Furthermore, the value of ImageNet pre-training \emph{seems to be} localized to very early layers and the internalization of domain-relevant data statistics \cite{Neyshabur2020, Raghu2019}. Future work could look into which of these may be leveraged with a new aggregate (multiple subjects \emph{and} tasks) pre-training, or the common subject-specific fine-tuning. This may provide insight into better weight initialization, or integration of explicit early layers similar to \cite{Raghu2019} (one could also argue that SincNet layers \cite{Ravanelli2018} are some such layers that could factor here). Additionally, as temporally-minded reconstruction losses continue to develop \cite{Rivest2020}, reconsidering the effectiveness of signal reconstruction as a pre-training objective (and/or regularization) is warranted, whether this is within an MLM-like scheme similar to BENDR, or a seq2seq model \cite{Graves2012}.

\section{Conclusion}
We have proposed MLM-like training as a self-supervised pre-training step for BCI/EEG DNNs. This is in the interest of diversifying the investigations into successful transfer learning schemes for DNNs applied to BCI and EEG. While previous approaches fashioned DNN transfer learning after ImageNet pre-training, we find this approach inadequate as there is limited applicable data availability and it is questionably analogous to its forebear. While our proposed alternative might similarly suffer from this latter point to some degree (the most distinct MLM success is with discrete sequences, not continuous ones), it is more conducive to leveraging potentially immense amounts of unlabelled data, it is not limited to long-term feature developments as with previous proposals, and it seems to produce representations equally suited to different users and sessions, which is a problem ImageNet pre-training appears less suited to solving. In summary, we see strong paths for the effective deployment of powerful computation and massive data scales with EEG and BCI. Effective solutions in these specific applications could help drive application \emph{and} analysis solutions in neuroimaging and perhaps physiology generally.

\bibliographystyle{unsrt}
\bibliography{references,hand_added}

\clearpage

\appendix

\section{Downstream hyperparameters}
\label{app:hyperparams}

\begin{table}[h]
    \centering
    \setlength{\extrarowheight}{4pt}
    \begin{tabular}{|l|c|c|c|c|c|c}
    \hline
        Dataset & Batch Size & Epochs & Learning Rate \\ \hline \hline
        MMI     & 4          & 7      & $1 \times 10^{-5}$\\ \hline
        BCIC    & 60         & 15     & $5 \times 10^{-5}$\\ \hline
        ERN     & 32         & 15     & $1 \times 10^{-5}$    \\ \hline
        P300    & 80         & 20     & $1 \times10^{-5}$\\ \hline
        SSC     & 64         & 40     & $5 \times 10^{-5}$\\ \hline
    \end{tabular}
    \caption{Hyperparameters that varied between datasets, these were not changed between different model configurations (see list in section \ref{sec:finetuning}).}
    \label{tab:ds-hyperparameters}
\end{table}

\end{document}